%% file: ML4eng_main.tex
\documentclass{article}

      \usepackage[nonatbib]{ml4eng_2020}

\usepackage[utf8]{inputenc} 
\usepackage[T1]{fontenc}    
\usepackage{hyperref}       
\usepackage{url}            
\usepackage{booktabs}       
\usepackage{amsfonts}       
\usepackage{nicefrac}       
\usepackage{microtype}      
\usepackage{times}
\usepackage{soul}
\usepackage{url}
\usepackage[small]{caption}
\usepackage{graphicx}
\usepackage{amsmath}
\usepackage{amsthm}
\usepackage{booktabs}
\usepackage{algorithmic}
\usepackage{algorithm}
\usepackage{wrapfig,lipsum,booktabs}
\usepackage{tcolorbox}
\newcommand\numberthis{\addtocounter{equation}{1}\tag{\theequation}}
\renewcommand{\vec}[1]{\mathbf{#1}}
\usepackage{multirow}
\numberwithin{equation}{section}
\usepackage{amssymb}
\title{Information-Theoretic Multi-Objective Bayesian Optimization with Continuous Approximations}

\author{%
   Syrine Belakaria,
   Aryan Deshwal,
  Janardhan Rao Doppa \\
  School of EECS, Washington State University\\
  \texttt{\{syrine.belakaria, aryan.deshwal, jana.doppa\}@wsu.edu}
}
\begin{document}
\setlength{\abovedisplayskip}{3pt}
\setlength{\belowdisplayskip}{3pt}
\maketitle
\begin{abstract}
Many real-world applications involve black-box optimization of multiple objectives using continuous function approximations that trade-off accuracy and resource cost of evaluation. For example, in rocket launching research, we need to find designs that trade-off return-time and angular distance using continuous-fidelity simulators (e.g., varying tolerance parameter to trade-off simulation time and accuracy) for design evaluations. The goal is to approximate the optimal Pareto set by minimizing the cost for evaluations.  In this paper, we propose a novel approach referred to as {\em {\bf i}nformation-Theoretic {\bf M}ulti-Objective Bayesian {\bf O}ptimization with {\bf C}ontinuous {\bf A}pproximations (iMOCA)} to solve this problem. The key idea is to select the sequence of input and function approximations for multiple objectives which maximize the information gain per unit cost for the optimal Pareto front. Our experiments on diverse synthetic and real-world benchmarks show that iMOCA significantly improves over existing single-fidelity methods.
\end{abstract}
\input{Introduction.tex}
\input{Related_work.tex}
\input{Problem_setup.tex}

\input{Solution.tex}
\input{Experiments.tex}
\section{Conclusions}
We introduced a novel approach referred as iMOCA to solve multi-objective Bayesian optimization problems with continuous function approximations. The key idea is to select inputs and function approximations for evaluation which maximizes the information gained per unit cost about the optimal Pareto front. Our experimental results on diverse benchmarks showed that iMOCA consistently outperforms state-of-the-art single-fidelity methods and a naive continuous-fidelity MO algorithm.

\vspace{2.0ex}

\noindent {\bf Acknowledgements.} The authors gratefully acknowledge the support from National Science Foundation (NSF) grants IIS-1845922 and OAC-1910213. The views expressed are those of the authors and do not reflect the official policy or position of the NSF.



\input{ML4eng_main.bbl}
\appendix
\input{appendix.tex}

\end{document}

%% file: Introduction.tex
\section{Introduction}
A large number of real-world engineering and scientific design problems involve optimizing {\em multiple} expensive black-box functions with {\em continuous approximations} (also called as fidelities) which vary in accuracy and resource cost. For example, in rocket launching research, the goal is to find designs that trade-off return time and angular distance using continuous-fidelity simulators which trade-off accuracy of evaluation and simulation time by varying the tolerance parameter. Solving these problems require finding the optimal Pareto set of design inputs (as all objectives cannot be optimized simultaneously) while minimizing the total resource cost of function evaluations.

For solving such expensive blackbox optimization problems, Bayesian optimization (BO) \cite{BO-Survey,BOCS,MOOS,PSR,DBO} is an effective framework. BO methods intelligently select the sequence of inputs for evaluation using the following three key elements: {\bf 1)} Surrogate statistical model learned from past function evaluations data, e.g., Gaussian process (GP); {\bf 2)} Acquisition function parametrized by the statistical model to score usefulness of evaluating different inputs, e.g., expected improvement (EI); and {\bf 3)} Acquisition function optimization to select the best input for function evaluation in each BO iteration. There are two broad categories of acquisition functions in the BO literature: {\em myopic} family accounts for local utility of improvement, e.g., EI; and {\em non-myopic} family relying on global utility for solving the optimization problem when evaluating a candidate input, e.g., predictive entropy search \cite{PES}. Recent work has shown the theoretical and practical benefits of non-myopic acquisition functions over myopic ones \cite{PES,hoffman2015output,MES}.

In this paper, we propose a novel approach referred as {\em {\bf i}nformation-Theoretic {\bf M}ulti-Objective Bayesian {\bf O}ptimization with {\bf C}ontinuous {\bf A}pproximations (iMOCA)} to solve multi-objective optimization problems using continuous-fidelity function evaluations. {\em To the best of our knowledge, this is the first work to study this continuous fidelity in the multi-objective setting within the ML literature}. The key idea behind iMOCA is to select the sequence of candidate input and fidelity-vector pairs for evaluation which maximize the information gained per unit cost about the optimal Pareto front (i.e., non-myopic acquisition function).  To efficiently compute entropy, which is an important step for iMOCA, we develop two qualitatively different approximations that make different trade-offs in terms of computation-time and accuracy.
iMOCA extends the single-fidelity multi-objective algorithm MESMO (\cite{belakaria2019max}) and its discrete-fidelity version (\cite{belakaria2020multi}) to the more general continuous-fidelity setting.

\vspace{1.0ex}
\noindent {\bf Contributions.} We make the following specific contributions.
\begin{itemize}
\setlength\itemsep{0em} 
\item Development of a principled information-theoretic approach referred as iMOCA to solve multi-objective blackbox optimization problems using continuous function approximations.  
Providing two qualitatively different approximations for iMOCA.
\item Experimental evaluation on diverse synthetic and real-world benchmark problems to demonstrate the effectiveness of iMOCA over existing single-fidelity MO algorithms and a naive continuous-fidelity MO approach. 
\end{itemize}

%% file: Related_work.tex
\section{Related work}
\noindent {\bf Multi-fidelity single-objective optimization.}

acquisition function for single-fidelity and single-objective BO has been extensively studied (\cite{BO-Survey}).  
Canonical examples of myopic acquisition function include expected improvement (EI) and upper-confidence bound (UCB). EI was extended to multi-fidelity setting (\cite{huang2006sequential,picheny2013quantile,lam2015multifidelity}). The popular GP-UCB method (\cite{gp-ucb}) was also extended to multi-fidelity setting with discrete fidelities (\cite{kandasamy2016gaussian}) and continuous fidelities (\cite{kandasamy2017multi}). 

Entropy based methods fall under the category of {\em non-myopic} acquisition function 
Some examples include entropy search (ES) (\cite{entropy_search}) and predictive entropy search (PES) (\cite{PES}). Their multi-fidelity extensions include MT-ES (\cite{swersky2013multi,klein2017fast}) and MF-PES (\cite{zhang2017information,mcleod2017practical}). Unfortunately, they inherit the computational difficulties of the original ES and PES. Max-value entropy search (MES) and output space predictive entropy search (\cite{MES,hoffman2015output}) are recent approaches that rely on the principle of output space entropy (OSE) search. Prior work (\cite{MES}) has shown advantages of OSE search in terms of compute-time, robustness, and accuracy over input space entropy search methods.
Recent work (\cite{song2018general}) proposed a general approach based on mutual information. \cite{takeno2019multi} extended MES to multi-fidelity setting and showed its effectiveness over MF-PES. \cite{moss2020mumbo} extended MES to the continuous fidelity and multi-task setting

\vspace{1.0ex}

\noindent {\bf Single-fidelity multi-objective optimization.} Multi-objective algorithms can be classified into three families. {\em Scalarization methods} are model-based algorithms that reduce the problem to single-objective optimization. ParEGO method (\cite{knowles2006parego}) employs random scalarization for this purpose. 
ParEGO is simple and fast, but more advanced approaches often outperform it. {\em Pareto hypervolume optimization methods} optimize the Pareto hypervolume (PHV) metric (\cite{emmerich2008computation}) that captures the quality of a candidate Pareto set. This is done by extending the standard acquisition functions to PHV objective, e.g., expected improvement in PHV (\cite{emmerich2008computation}) and probability of improvement in PHV (\cite{picheny2015multiobjective}). Unfortunately, algorithms to optimize PHV based acquisition functions scale very poorly and are not feasible for more than three objectives. To improve scalability, methods to reduce the search space are also explored  (\cite{ponweiser2008multiobjective}). A common drawback of this family is that reduction to single-objective optimization 
can potentially lead to more exploitative behavior.

{\em Uncertainty reduction methods} like PAL (\cite{zuluaga2013active}), PESMO (\cite{PESMO}) and the concurrent works USeMO (\cite{Usemo}) and MESMO (\cite{belakaria2019max}) are principled algorithms based on information theory. 
In each iteration, PAL selects the candidate input for evaluation towards the goal of minimizing the size of uncertain set. PAL provides theoretical guarantees, but it is only applicable for input space $\mathcal{X}$ with finite set of discrete points. USeMO is a general framework that iteratively generates a cheap Pareto front using the surrogate models and then selects the point with highest uncertainty as the next query. USeMOC \cite{USeMOC} is an extension of USeMO for handling constraints and was applied to design analog circuits \cite{DATE-2020}.
PESMO relies on input space entropy
search and iteratively selects the input that maximizes the information gained about the optimal Pareto set $\mathcal{X}^*$. Unfortunately, optimizing this acquisition function poses significant challenges: a) requires a series of approximations, which can be potentially sub-optimal; and b) optimization, even after approximations, is expensive c) performance is strongly dependent on the number of Monte-Carlo samples. MESMO (\cite{belakaria2019max}) relies on output space entropy search and its advantages over PESMO were demonstrated in a recent work \cite{belakaria2019max}. MESMOC \cite{MESMOC} is an extension of MESMO for handling constraints and was applied to design electrified aviation power systems \cite{belakaria2020PSD}.

\vspace{1.0ex}

\noindent {\bf Multi-fidelity multi-objective optimization.} Prior work outside ML literature has considered domain-specific methods that employ single-fidelity multi-objective approaches in the context of multi-fidelity setting by using the lower fidelities {\em only as an initialization} (\cite{kontogiannis2018comparison,ariyarit2017multi}). Specifically, (\cite{ariyarit2017multi}) employs the single-fidelity algorithm based on expected hypervolume improvement acquisition function and (\cite{kontogiannis2018comparison}) employs an algorithm that is very similar to SMSego. Additionally, both these methods model all fidelities with the same GP and assume that higher fidelity evaluation is a sum of lower-fidelity evaluation and offset error. These are strong assumptions and may not hold in general multi-fidelity settings including the problems we considered in our experimental evaluation. A recent work \cite{belakaria2020multi} proposed a generic approach by generalizing the MESMO algorithm \cite{belakaria2019max} for the discrete-fidelity setting. However, this approach assumes the existence of different functions for each fidelity which makes it hard to extend/apply to the continuous-fidelity setting. We are not aware of any continuous-fidelity algorithms for MO problems.

%% file: Problem_setup.tex
\section{Problem Setup}
{\bf Multi-objective optimization with continuous function approximations.} Suppose $\mathcal{X} \subseteq \Re^d$ is a continuous input space. In multi-objective optimization (MO) problems, our goal is to minimize $K \geq 2$ {\em expensive} objective functions $f_1(\vec{x}), f_2(\vec{x}),\cdots,f_K(\vec{x})$. The evaluation of each input $\vec{x}\in \mathcal{X}$  results in a vector of $K$ function values $\vec{y}$ = $(y_1, y_2,\cdots,y_K)$, where $y_i$ = $f_i(x)$ for all $i \in \{1,2, \cdots, K\}$.  An input $\vec{x} \in \mathcal{X}$ is said to {\em Pareto-dominate} another input  $\vec{x'} \in \mathcal{X}$ if $f_i(\vec{x}) \leq f_i(\vec{x'}) \hspace{1mm} \forall{i}$ and there exists some $j \in \{1, 2, \cdots,K\}$ such that $f_j(\vec{x}) < f_j(\vec{x'})$. The solution for this MO problem referred as optimal {\em Pareto set} $\mathcal{X}^* \subset \mathcal{X}$ is a set of non-dominated inputs  and optimal {\em Pareto front} 
is the corresponding set of function value vectors. 

In continuous-fidelity MO problems, we have access to $g_i(\vec{x},z_i)$ where $g_i$ is an alternative function through which we can evaluate cheaper approximations of $f_i$ by varying the fidelity variable $z_i \in \mathcal{Z}$. Without loss of generality, let $\mathcal{Z}$=$\left[0, 1 \right]$ be the fidelity space. Fidelities for each function $f_i$ vary in the amount of resources consumed and the accuracy of evaluation, where $z_i$=0 and $z_i^*$=1 refer to the lowest and highest fidelity respectively. At the highest fidelity $z_i^*$, $g_i(\vec{x},z_i^{*})=f_i(\vec{x})$. 
 Let $\mathcal{C}_i(\vec{x},z_i)$ be the cost of evaluating $g_i(\vec{x},z_i)$. Evaluation of an input $\vec{x}\in \mathcal{X}$ with fidelity vector $\vec{z} = [z_1, z_2, \cdots, z_K]$ produces an evaluation vector of $K$ values denoted by $\vec{y} \equiv [y_1, y_2,\cdots, y_K]$, where $y_i = g_i(\vec{x},z_i)$ for all $i \in \{1,2, \cdots, K\}$, and the normalized cost of evaluation is $\mathcal{C}(\vec{x},\vec{z}) = \sum_{i=1}^{K} \left( {\mathcal{C}_i(\vec{x},z_i)}/{\mathcal{C}_i(\vec{x},z_i^*)}\right)$. We normalize the cost of each function by the cost of its highest fidelity because the cost units of different objectives can be different.
 If the cost is known, it can be directly injected in the latter expression. However, in some real wold setting, the cost of a function evaluation can be only known after the function evaluation. For example, in hyper-parameter tuning of a neural network, the cost of the experiment is defined by the training and inference time, However, we cannot know the exact needed time until after the experiment is finalised. In this case, the cost can be modeled by an independent Gaussian process. The predictive mean can be used during the optimization.  The final goal is to recover $\mathcal{X}^*$ while minimizing the total 
 cost of function evaluations.
 
 \vspace{1.0ex}
 
\noindent {\bf Continuous-fidelity GPs as surrogate models.} \label{surrogatesection}
Let $D$ = $\{(\vec{x}_i, \vec{y}_i,\vec{z}_i)\}_{i=1}^{t-1}$ be the training data from past $t$-1 function evaluations, where  $\vec{x}_i \in \mathcal{X}$ is an input and $\vec{y}_i = [y_1,y_2,\cdots,y_K]$ is the output vector resulting from evaluating functions $g_1, g_2,\cdots,g_K$ for $\vec{x}_i$ at fidelities $z_1, z_2,\cdots,z_K$ respectively. We learn $K$ surrogate statistical models $\mathcal{GP}_1,\cdots,\mathcal{GP}_K$ from $\mathcal{D}$, where each model $\mathcal{GP}_j$ corresponds to the $j$th function $g_j$. 
Continuous fidelity GPs (CF-GPs) are capable of modeling functions with continuous fidelities within a single model. Hence, we employ CF-GPs to build surrogate statistical models for each function. Specifically, we use the CF-GP model proposed in \cite{kandasamy2017multi}.
W.l.o.g, we assume that our functions $g_j$ are defined over the spaces $\mathcal{X}=[0,1]^d$ and $\mathcal{Z}=[0,1]$. Let $g_j \sim \mathcal{GP}_j(0,\kappa_j)$ such that $y_j=g_j(z_j,\vec{x})+\epsilon$, where $\epsilon \sim \mathcal{N}(0, \eta^2)$ and $\kappa: (\mathcal{Z} \times \mathcal{X})^2 \rightarrow \mathbb{R} $ is the prior covariance matrix defined on the product of input and fidelity spaces. 
\begin{align*}
    \kappa_j([z_j,\vec{x}],[z_j',\vec{x}'])= \kappa_{j\mathcal{X}}(\vec{x},\vec{x}') \cdot \kappa_{j\mathcal{Z}}(z_j,z_j')
\end{align*}
where $\kappa_{j\mathcal{X}},\kappa_{j\mathcal{Z}}$ are radial kernels over $\mathcal{X}$ and $\mathcal{Z}$ spaces respectively. $\mathcal{Z}$ controls the smoothness of $g_j$ over the fidelity space to be able to share information across different fidelities. 
A key advantage of this model is that it integrates all fidelities into one single GP for information sharing. 
We denote the posterior mean and standard deviation of $g_j$ conditioned on $D$ by $\mu_{g_j}(\vec{x},z_j)$ and $\sigma_{g_j}(\vec{x},z_j)$. We denote the posterior mean and standard deviation of the highest fidelity functions $f_j(\vec{x})=g_j(\vec{x},z_j^*)$ by $\mu_{f_j}(x)=\mu_{g_j}(\vec{x},z_j^*)$ and $\sigma_{f_j}(\vec{x})=\sigma_{g_j}(\vec{x},z_j^*)$ respectively. We define $\sigma_{g_j,f_j}^2(x)$ as the predictive co-variance between a lower fidelity $z_j$ and the highest fidelity $z_j^*$ at the same $\vec{x}$.

\vspace{1.0ex}

\noindent{\bf Table of Notations.} For the sake of reader, Table~\ref{table:notations} provides all the mathematical notations used in this paper.\\

\begin{table*}[h!]
    \centering
    \resizebox{1\linewidth}{!}{
    \begin{tabular}{|c|l|}
    \hline
    {\bf Notation} & {\bf Definition} \\
    \hline
       $\vec{x}, \vec{y}, \vec{z}$ & Bold notation represents vectors\\
       \hline
       $f_1, f_2, \cdots, f_K$  & Highest fidelity objective functions \\
       \hline
              $g_1, g_2, \cdots, g_K$  & General objective functions with low and high fidelities\\
       \hline
$\Tilde{f}_j$ & Function sampled from the highest fidelity of the $j$th Gaussian process model\\
\hline
$\Tilde{g}_j$ & Function sampled from the $j$th Gaussian process model at fidelity $z_j$\\
       \hline
       $\vec{x}$ & Input vector \\
              \hline
       $z_1, z_2, \cdots, z_K$ & The fidelity variables for each function \\
       \hline

       $\vec{z}$ & Fidelity vector \\
       \hline
         $\vec{z}^*=[z_1^*, z_2^*, \cdots, z_K^*]$ & Fidelity vector with all fidelities at their highest value\\
       \hline
        $y_j$ &$j$th function $g_j$ evaluated at fidelity $z_j$  \\
       \hline
      $\vec{y}=[y_1,y_2,\cdots,y_K]$   & Output vector resulting from evaluating $g_1, g_2,\cdots,g_K$ \\ 
      & for $\vec{x}_i$ at fidelities $z_1, z_2,\cdots,z_K$ respectively\\  
      \hline 
            $\vec{f}=[f_1,f_2,\cdots,f_K]$   & Output vector resulting from evaluating functions $f_1, f_2,\cdots,f_K$ \\
            & or equivalently $g_1, g_2,\cdots,g_K$ for $\vec{x}_i$ at fidelities $z^*_1, z^*_2,\cdots,z^*_K$ respectively\\  
      \hline 
$\mathcal{F}^*$ & true Pareto front of the highest fidelity functions $[f_1, f_2, \cdots, f_K]$\\ 
\hline
$\mathcal{F}_s^*$ &  Pareto front of the sampled highest fidelity functions $[\Tilde{f}_1, \Tilde{f}_2, \cdots, \Tilde{f}_K]$\\ 
\hline 
$\mathcal{C}_j(\vec{x},z_j)$ & cost of evaluating $j$th function $g_j$ at fidelity $z_j$\\
\hline
$\mathcal{C}(\vec{x},\vec{z})$ & total normalized cost $\mathcal{C}(\vec{x},\vec{z}) = \sum_{i=1}^{K} \left({\mathcal{C}_i(\vec{x},z_i)}/{\mathcal{C}_i(\vec{x},z_i^*)}\right)$  \\
\hline
$\mathcal{X}$& Input space \\
\hline 
$\mathcal{Z}$ & Fidelity space \\
\hline
$\mathcal{Z}_t^{(j)}$& Reduced fidelity space for function $g_j$ at iteration $t$ \\
\hline
$\mathcal{Z}_r$& Reduced fidelity space over all the functions\\
\hline
$\xi$ & Information gap \\
\hline
$\beta_t^{(j)}$ &  Exploration/exploitation parameter for function $g_j$ at iteration $t$ \\
\hline
$I$ & Information gain \\
\hline

\hline

    \end{tabular}}
    \caption{Mathematical notations and their associated definition.}
    \label{table:notations}
\end{table*}

%% file: Solution.tex
\section{iMOCA Algorithm with Two Approximations}
We first describe the key idea behind our proposed iMOCA algorithm including the main challenges. Next, we present our  algorithmic solution to address those challenges.

\vspace{1.0ex}

\noindent {\bf Key Idea of iMOCA:} The acquisition function behind iMOCA employs principle of output space entropy search to select the sequence of input and fidelity-vector (one for each objective) pairs. iMOCA is applicable for solving MO problems in both continuous and discrete fidelity settings. {\em The key idea is to find the pair $\{\vec{x}_t, \vec{z}_t\}$ that maximizes the information gain $I$ per unit cost about the Pareto front of the highest fidelities} (denoted by $\mathcal{F}^*$), where $\{\vec{x}_t, \vec{z}_t\}$ represents a candidate input $\vec{x}_t$ evaluated at a vector of fidelities $\vec{z}_t$ = $[z_1, z_2, \cdots, z_K]$ at iteration $t$. Importantly, iMOCA performs joint search over input space $\mathcal{X}$ and reduced fidelity space $\mathcal{Z}_r$ over fidelity vectors for this selection.
\begin{align}
    (\vec{x}_{t},\vec{z}_t) \leftarrow \arg max_{\vec{x}\in \mathcal{X},\vec{z}\in \mathcal{Z}_r}\hspace{2 mm}\alpha_t(\vec{x},\vec{z}) ~
        , \; \text{where} \quad \alpha_t(\vec{x},\vec{z}) &= I(\{\vec{x}, \vec{y},\vec{z}\}, \mathcal{F}^* | D) / \mathcal{C}(\vec{x},\vec{z}) \label{af:def}
\end{align}
In the following sections, we describe the details and steps of our proposed algorithm iMOCA. We start by explaining the bottlenecks of continuous fidelity optimization due to the infinite size of the fidelity space followed 
by 
describing a principled approach to reduce the fidelity space. Subsequently, we present the computational steps of our proposed acquisition function: Information gain per unit cost for each candidate input and fidelity-vector pair. 

\subsection{Approach to Reduce Fidelity Search Space}
In this work, we focus primarily on MO problems with continuous fidelity space. The continuity of this space results in infinite number of fidelity choices. Thus, selecting an informative and meaningful fidelity becomes a major bottleneck. Therefore, we reduce the search space over fidelity-vector variables in a principled manner guided by the learned statistical models \cite{kandasamy2017multi}. Our fidelity space reduction method is inspired from BOCA for single-objective optimization \cite{kandasamy2017multi}. We apply the method in BOCA to each of the objective functions to be optimized in MO setting.

A favourable setting for continuous-fidelity methods would
be for the lower fidelities $g_j$ to be informative about the highest fidelity $f_j$. Let $h_j$ be the bandwith parameter of the fidelity kernel $\kappa_{j\mathcal{Z}}$ and let $\xi:\mathcal{Z}\rightarrow [0,1]$ be a measure of the gap in information about $g_j(.,z_j^*)$ when queried at $z_j\neq z_j^*$ with $\xi(z_j)\approx \frac{\Vert z_j - z_j^* \Vert}{h_j}$ for the squared exponential kernels \cite{kandasamy2017multi}. 
A larger $h_j$ will result in $g_j$ being smoother across $\mathcal{Z}$. Consequently, lower fidelities will be more informative about $f_j$ and the information gap $\xi(z_j)$ will be smaller.  

To determine an informative fidelity for each function in iteration $t$, we reduce the space $\mathcal{Z}$ and select $z_j$ from the subset $ \mathcal{Z}_t^{(j)}$ defined as follows:
\begin{align*}
    \mathcal{Z}_t^{(j)}(\vec{x})=\{\{z_j \in  \mathcal{Z}_{\backslash \{z_j^*\}}, \sigma_{g_j}(\vec{x},z_j) > \gamma(z_j), \xi(z_j) > \beta_t^{(j)} \Vert \xi \Vert_{\infty} \} \cup \{z_j^*\}\} \numberthis \label{spacereduce}
\end{align*} 
where $\gamma(z_j)=\xi(z_j) (\frac{\mathcal{C}_j(\vec{x},z_j)}{\mathcal{C}_j(\vec{x},z_j^*)})^q$ and $q=\frac{1}{p_j+d+2}$ with $p_j,d$ the dimensions of $\mathcal{Z}$ and $\mathcal{X}$ respectively. Without loss of generality, we assume that $p_j=1$.
$\beta^{(j)}_t = \sqrt{0.5 \ln(\frac{2t+1}{h_j})}$ is the exploration/exploitation parameter \cite{gp-ucb}.
We denote by $\mathcal{Z}_r=\{\mathcal{Z}_t^{(j)} , j \in \{1 \dots K\}\}$, the reduced fidelity space for all $K$ functions. We filter out the fidelities for each objective function at BO iteration $t$ using the above-mentioned two conditions. We provide intuitive explanation of these conditions below. 

\noindent \textbf{The first condition $\sigma_{g_j}(\vec{x},z_j) > \gamma(z_j)$:} A reasonable multi-fidelity strategy would query the cheaper fidelities in the beginning to learn about the function $g_j$ by consuming the least possible cost budget and later query from higher fidelities in order to gain more accurate information. Since the final goal is to optimize $f_j$, the algorithm should also query from the highest-fidelity. However, the algorithm might never query from higher fidelities due to their high cost. 
This condition will make sure that lower fidelities are likely to be queried, but not excessively and the algorithm will move toward querying higher fidelities as iterations progress. Since $\gamma(z_j)$ is monotonically
increasing in $\mathcal{C}_j$, this condition can be easily satisfied by cheap fidelities. However, if a fidelity is very far from $z_j^*$, then the information gap $\xi$ will increase and hence, uninformative fidelities would be discarded. Therefore, $\gamma(z_j)$ will guarantee achieving a good trade-off between resource cost and information.

\noindent \textbf{The second condition $\xi(z_j) > \beta_t^{(j)}\Vert \xi \Vert_{\infty}$:} We recall that if the first subset of $\mathcal{Z}_t^{(j)}$ is empty, the algorithm will automatically evaluate the highest-fidelity $z_j^*$. However, if it is not empty, and since the fidelity space is continuous (infinite number of choices for $z_j$), the algorithm might query fidelities that are very close to $z_j^*$ and would cost nearly the same as $z_j^*$ without being as informative as $z_j^*$. The goal of this condition is to prevent such situations by excluding fidelities in the small neighborhood of $z_j^*$ and querying $z_j^*$ instead. Since $\beta_t^{(j)}$ increases with $t$ and $\xi$ is increasing as we move away from $z_j^*$, this neighborhood is shrinking and the algorithm will eventually query $z_j^*$.

\subsubsection{Naive-CFMO: A Simple Continuous-Fidelity MO Baseline}\label{cf-naive-section}
In this section, we first describe a simple baseline approach referred as {\em Naive-CFMO} to solve continuous-fidelity MO problems by combining the above-mentioned fidelity space reduction approach with existing multi-objective BO methods. Next, we explain the key drawbacks of Naive-CFMO and how our proposed iMOCA algorithm overcomes them. A straightforward way to construct a continuous-fidelity MO method is to perform a two step selection process similar to the continuous-fidelity single-objective BO algorithm proposed in \cite{kandasamy2017multi}: 

\vspace{0.5ex}

\noindent {\bf Step 1)} Select the input $\vec{x}$ that maximizes the acquisition function at the {\em highest fidelity}. This can be done using any existing multi-objective BO algorithm.

\vspace{0.5ex}

\noindent {\bf Step 2)} Evaluate $\vec{x}$ at the {\em cheapest valid fidelity for each function} in the reduced fidelity space $ \mathcal{Z}_t^{(j)}(\vec{x})$ computed using the reduction approach mentioned in the previous section. Since we are studying information gain based methods in this work, we instantiate Naive-CFMO using the state-of-the-art information-theoretic MESMO algorithm \cite{belakaria2019max} for Step 1. Algorithm \ref{alg:CFMESMONaive} shows the complete pseudo-code of Naive-CFMO. 

\vspace{1.0ex}

\noindent {\bf Drawbacks of Naive-CFMO:} Unfortunately, Naive-CFMO has two major drawbacks.
\begin{itemize}

\item The acquisition function solely relies on the highest-fidelity $f_j$. Therefore, it does not capture and leverage the statistical relation between  different fidelities and full-information provided by the global function $g_j$. 

\item Generally, there is a dependency between the fidelity space and the input space in continuous-fidelity problems. Therefore, selecting an input that maximizes the highest-fidelity and then evaluating it at a different fidelity can result in a mismatch in the evaluation process leading to poor performance and slower convergence.
\end{itemize}

\vspace{1.0ex}

\noindent {\bf iMOCA vs. Naive-CFMO:} Our proposed iMOCA algorithm overcomes the drawbacks of Naive-CFMO as follows. 

\begin{itemize}

\item  iMOCA's acquisition function maximizes the information gain per unit cost across all fidelities by capturing the relation between fidelities and the impact of resource cost on information gain. 

\item iMOCA performs joint search over input and fidelity space to select the input variable $\vec{x} \in \mathcal{X}$ and fidelity variables $\vec{z} \in \mathcal{Z}_r$ while maximizing the proposed acquisition function. Indeed, our experimental results demonstrate the advantages of iMOCA over Naive-CFMO.

\end{itemize}

\subsection{Information-Theoretic Continuous-Fidelity Acquisition Function} 
In this section, we explain the technical details of the acquisition function behind iMOCA. We propose two approximations for the computation of information gain per unit cost.

The information gain in equation \ref{af:def} is defined as the expected reduction in entropy $H(.)$ of the posterior distribution $P(\mathcal{F}^* | D)$ due to evaluating $\vec{x}$ at fidelity vector $\vec{z}$. Based on the symmetric property of information gain, the latter can be rewritten as follows:

\begin{align}
    I(\{\vec{x}, \vec{y},\vec{z}\}, \mathcal{F}^{*} | D) &= H(\vec{y} | D, \vec{x},\vec{z}) - \mathbb{E}_{\mathcal{F}^{*}} [H(\vec{y} | D,   \vec{x},\vec{z}, \mathcal{F}^{*})] \numberthis \label{eqn_symmetric_ig}
\end{align}
 In equation \ref{eqn_symmetric_ig}, the first term is the entropy of a $K$-dimensional Gaussian distribution  that can be computed in closed form as follows:
\begin{align}
H(\vec{y} | D, \vec{x},\vec{z}) = \sum_{j = 1}^K \ln (\sqrt{2\pi e} ~ \sigma_{g_j}(\vec{x},z_j)) \numberthis \label{firstpart}
\end{align}
In equation \ref{eqn_symmetric_ig}, the second term is an expectation over the Pareto front of the highest fidelities $\mathcal{F}^{*}$. This term can be approximated using Monte-Carlo sampling: 
\begin{align}
    \mathbb{E}_{\mathcal{F}^{*}} [H(\vec{y} | D,   \vec{x},\vec{z}, \mathcal{F}^{*})] \simeq \frac{1}{S} \sum_{s = 1}^S [H(\vec{y} | D,   \vec{x},\vec{z}, \mathcal{F}^{*}_s)] \label{eqn_summation}
\end{align}
where $S$ is the number of samples and $\mathcal{F}^{*}_s$ denote a sample Pareto front obtained over the highest fidelity functions sampled  from $K$ surrogate models.  
To compute Equation \ref{eqn_summation}, we provide algorithmic solutions to construct Pareto front samples $\mathcal{F}^{*}_s$ and to compute the entropy with respect to a given Pareto front sample $\mathcal{F}^{*}_s$. 

\vspace{1.0ex}

\noindent {\bf Computing Pareto front samples:} We first sample highest fidelity functions $\Tilde{f}_1,\cdots,\Tilde{f}_K$ from the posterior CF-GP models via random fourier features \cite{PES,random_fourier_features}. This is done similar to prior work \cite{PESMO,MES}. We solve a cheap MO optimization problem over the $K$ sampled functions $\Tilde{f}_1,\cdots,\Tilde{f}_K$ using the popular NSGA-II algorithm \cite{deb2002nsga} to compute the sample Pareto front $\mathcal{F}^{*}_s$. 

\vspace{1.0ex}

\noindent {\bf Entropy computation for a given Pareto front sample:} Let $\mathcal{F}^{*}_s = \{\vec{v}^1, \cdots, \vec{v}^l \}$ be the sample Pareto front, where $l$ is the size of the Pareto front and each $\vec{v}^i = \{v_1^i,\cdots,v_K^i\}$ is a $K$-vector evaluated at the $K$ sampled highest-fidelity functions. The following inequality holds for each component $y_j$ of $\vec{y}$ = $(y_1, \cdots, y_K)$ in the entropy term $H(\vec{y} | D,   \vec{x},\vec{z}, \mathcal{F}^{*}_s)$:
\begin{align}
 y_j &\leq f_s^{j*} \quad \forall j \in \{1,\cdots,K\} \label{inequality}
\end{align}
where $f_s^{j*} = \max \{v_j^1, \cdots v_j^l \}$. Essentially, this inequality says that the $j^{th}$ component of $\vec{y}$ (i.e., $y_j$) is upper-bounded by a value, which is the maximum of $j^{th}$ components of all $l$ vectors $\{\vec{v}^1, \cdots, \vec{v}^l \}$ in the Pareto front $\mathcal{F}^{*}_s$.
{\em For the ease of notation, we drop the dependency on $\vec{x}$ and $\vec{z}$. We use $f_j$ to denote $f_j(x)=g_j(x,z_j^*)$, the evaluation of the highest fidelity at $x$ and $y_j$ to denote $g_j(x,z_j)$ the evaluation of $g_j$ at a lower fidelity $z_j \neq z_j^*$.} 

The proof of \ref{inequality} can be divided into two cases: {\bf a)} If $y_j$ is evaluated at its highest fidelity (i.e, $z_j=z_j^*$ and $y_j=f_j$), we provide a proof by contradiction for inequality \ref{inequality}. Suppose there exists some component $f_j$ of $\vec{f}$ such that $f_j > f_s^{j*}$. However, by definition, $\vec{f}$ is a non-dominated point because no point dominates it in the $j$th dimension. This results in $\vec{f} \in \mathcal{F}^*_s$ which is a contradiction. Therefore, inequality \ref{inequality} holds. {\bf b)} If $y_j$ is evaluated at one of its lower fidelities (i.e, $z_j \neq z_j^*$), the proof follows from the assumption that the value of lower fidelity of an objective is usually smaller than the corresponding higher fidelity, i.e., $y_j \leq f_j \leq f_s^{j*}$. This is especially true for real-world optimization problems. For example, in optimizing the accuracy of a neural network with respect to its hyper-parameters, a commonly employed fidelity is the number of data samples used for training. It is reasonable to believe that the accuracy is always higher for the higher fidelity. 

By combining the inequality \ref{inequality} and the fact that each function is modeled as an independent CF-GP, a common property of entropy measure allows us to decompose the entropy of a set of independent variables into a sum over entropies of individual variables \cite{information_theory}:
\begin{align}
H(\vec{y} | D,   \vec{x},\vec{z}, \mathcal{F}^{*}_s) \simeq \sum_{j=1}^K H(y_j|D, \vec{x},z_j,f_s^{j*}) \label{eqn_sep_ineq}
\end{align} 
The computation of \ref{eqn_sep_ineq} requires the computation of the entropy of $p(y_j|D, \vec{x},z_j,f_s^{j*})$. This is a conditional distribution that depends on the value of $z_j$ and can be expressed as $H(y_j|D, \vec{x},z_j,y_j\leq f_s^{j*})$. This entropy can be computed using two different approximations as described below:

\textbf{Truncated Gaussian approximation (iMOCA-T): }
As a consequence of \ref{inequality}, which states that $y_j\leq f_s^{j*}$ also holds for all fidelities, the entropy of $p(y_j|D, \vec{x},z_j,f_s^{j*})$ can also be approximated by the entropy of a truncated Gaussian distribution and expressed as follows:
\begin{align}
&H(y_j|D, \vec{x},z_j,y_j\leq f_s^{j*})=  \ln(\sqrt{2\pi e} ~ \sigma_{g_j}) +  \ln \Phi(\gamma_s^{(g_j)})- \frac{\gamma_s^{(g_j)} \phi(\gamma_s^{(g_j)})}{2\Phi(\gamma_s^{(g_j)})} ~ \text{where} ~ \gamma_s^{(g_j)} = \frac{f_s^{j*} - \mu_{g_j}}{\sigma_{g_j}}  \numberthis \label{entropyapprox1}
\end{align}
From equations \ref{eqn_summation}, \ref{firstpart}, and \ref{entropyapprox1}, we get the final expression of iMOCA-T as shown below:
\begin{align}
    \alpha_t(\vec{x},\vec{z},\mathcal{F}^{*})=&\frac{1}{\mathcal{C}(\vec{x},\vec{z}) S}\sum_{j=1}^K \sum_{s=1}^S  \frac{\gamma_s^{(g_j)}\phi(\gamma_s^{(g_j)})}{2\Phi(\gamma_s^{(g_j)})} - \ln(\Phi(\gamma_s^{(g_j)})) \numberthis \label{Tappriximation}
\end{align}

\textbf{Extended-skew Gaussian approximation (iMOCA-E):} Although equation \ref{Tappriximation} is sufficient for computing the entropy, this entropy can be mathematically interpreted and computed with a different approximation. The condition $y_j \leq f_s^{j*}$, is originally expressed as $f_j \leq f_s^{j*}$. Substituting this condition with it's original equivalent, the entropy becomes $H(y_j|D, \vec{x},z_j,f_j\leq f_s^{j*}) $. Since $y_j$ is an evaluation of the function $g_j$ while $f_j$ is an evaluation of the function $f_j$, we observe that $y_j | f_j \leq f_s^{j*}$ follows an extended-skew Gaussian (ESG) distribution \cite{moss2020mumbo,azzalini1985class}. It has been shown that the differential entropy of an ESG does not have a closed form expression \cite{arellano2013shannon}. Therefore, we derive a simplified expression where most of the terms are analytical by manipulating the components of the entropy. We use the derivation of the entropy based on ESG formulation, proposed in (\cite{moss2020mumbo}), for the multi-objective setting. 

In order to simplify the calculation $H(y_j|D, \vec{x},z_j,f_j\leq f_s^{j*})$,
let us define the normalized variable $\Gamma_{f_s^{j*}}$ as $\Gamma_{f_s^{j*}}= \frac{y_j - \mu_{g_j}}{\gamma_{g_j}} |f_j\leq f_s^{j*} $. $\Gamma_{f_s^{j*}}$ is an ESG with p.d.f whose mean $\mu_{\Gamma_{f_s^{j*}}}$ and variance $\sigma_{\Gamma_{f_s^{j*}}}$ are defined in Appendix A. We define the predictive correlation between $y_j$ and $f_j$ as $\tau=\frac{\sigma_{g_j,f_j}^2}{\sigma_{g_j}\sigma_{f_j}}$. 
The entropy can be computed using the following expression. Due to lack of space, we only provide the final expression. Complete derivation for equations \ref{entropyapprox2} and \ref{Eappriximation} are provided in Appendix A.
\begin{align}
    H(y_j|D, \vec{x},z_j,f_j\leq f_s^{j*}) &= \ln(\sqrt{2\pi e} ~\sigma_{g_j}) +\ln(\Phi(\gamma_s^{(f_j)})) - \tau^2\frac{\phi(\gamma_s^{(f_j)})\gamma_s^{(f_j)}}{2\Phi(\gamma_s^{(f_j)})}\nonumber \\
    & \quad - \mathbb{E}_{u \sim \Gamma_{f_s^{j*}}}\left[   \ln(\Phi(\frac{\gamma_s^{(f_j)}-\tau u}{\sqrt{1-\tau^2}}))\right] \numberthis \label{entropyapprox2}
\end{align}
From equations \ref{eqn_summation}, \ref{firstpart} and \ref{entropyapprox2}, the final expression of iMOCA-E can be expressed as follow:
\begin{align}
    \alpha_t(\vec{x},\vec{z},\mathcal{F}^{*})=&\frac{1}{\mathcal{C}(\vec{x},\vec{z})S}\sum_{j=1}^K \sum_{s=1}^S \tau^2\frac{\gamma_s^{(f_j)}\phi(\gamma_s^{(f_j)})}{2\Phi(\gamma_s^{(f_j)})} - \ln(\Phi(\gamma_s^{(f_j)})) +\mathbb{E}_{u \sim \Gamma_{f_s^{j*}}}[\ln(\Phi(\frac{\gamma_s^{(f_j)}-\tau u}{\sqrt{1-\tau^2}}))] \numberthis \label{Eappriximation}
\end{align}
The expression given by equation \ref{Eappriximation} is mostly analytical except for the last term. We perform numerical integration via Simpson’s rule using $\mu_{\Gamma_{f_s^{j*}}} \mp \gamma \sqrt{\sigma(\Gamma_{f_s^{j*}})} $ as the integral limits. Since this integral is over one-dimension variable, numerical integration can result in a tight approximation with low computational cost. Complete pseudo-code of  iMOCA is shown in Algorithm \ref{alg:CFMESMO}.

\vspace{1.0ex}

\textbf{Generality of the two approximations:} We observe that for any fixed value of $\vec{x}$, when we choose the highest-fidelity for each function $\vec{z}$=$\vec{z}^*$: {\bf a)} For iMOCA-T, we will have $g_i=f_j$; and {\bf b)} For iMOCA-E, we will have $\tau = 1$. Consequently, both equation \ref{Tappriximation} and  equation \ref{Eappriximation} will degenerate to the acquisition function of MESMO optimizing only highest-fidelity functions:
\begin{align}
   \alpha_t(\vec{x},\mathcal{F}^{*})= \frac{1}{S}\sum_{j=1}^K \sum_{s=1}^S \frac{\gamma_s^{(f_j)}\phi(\gamma_s^{(f_j)})}{2\Phi(\gamma_s^{(f_j)})} - \ln(\Phi(\gamma_s^{(f_j)})) \numberthis \label{MESMO}
\end{align}
 
The main advantages of our proposed acquisition function are: cost-efficiency, computational-efficiency, and robustness to the number of Monte-Carlo samples. Indeed, our experiments demonstrate these advantages over state-of-the-art single-fidelity MO algorithms.

\noindent
\begin{minipage}{0.49\textwidth}
\begin{algorithm}[H]
\centering
\scriptsize
\caption{iMOCA Algorithm}
\textbf{Input}: input space $\mathcal{X}$; $K$ blackbox functions $f_j$ and their continuous approximations $g_j$; total budget $\mathcal{C}_{total}$ 
\begin{algorithmic}[1] 
\STATE Initialize continuous fidelity gaussian process $\mathcal{GP}_1, \cdots, \mathcal{GP}_K$ by initial points $D$
\STATE \textbf{While} {$\mathcal{C}_{t} \leq \mathcal{C}_{total}$} \textbf{do}
 \STATE \quad for each sample $s \in {1,\cdots,S}$: 
 \STATE \quad \quad Sample highest-fidelity functions $\Tilde{f}_j \sim \mathcal{GP}_j(.,z_j^*) $
 \STATE \quad \quad $\mathcal{F}_s^{*} \leftarrow$ Solve {\em cheap} MOO over $(\Tilde{f}_1, \cdots, \Tilde{f}_K)$
 \STATE \quad Find the query based on $\mathcal{F}^{*}=\{\mathcal{F}_s^{*}, s \in \{1 \dots S\}\}$
 \textcolor{red}{\STATE \quad // Choose one of the two approximations
\STATE \quad {\bf If} approx = T // Use eq \ref{Tappriximation} for $\alpha_t$ (iMOCA-T)
\STATE \quad \quad select $(\vec{x}_{t},\vec{z}_t) \leftarrow \arg max_{\vec{x}\in \mathcal{X},\vec{z}\in \mathcal{Z}_r} \hspace{2 mm} \alpha_t(\vec{x},\vec{z},\mathcal{F}^{*})$
\STATE \quad {\bf If} approx = E // Use eq \ref{Eappriximation} for $\alpha_t$ (iMOCA-E)
\STATE \quad\quad select $(\vec{x}_{t},\vec{z}_t) \leftarrow \arg max_{\vec{x}\in \mathcal{X},\vec{z}\in \mathcal{Z}_r} \hspace{2 mm}\alpha_t(\vec{x},\vec{z},\mathcal{F}^{*})$ }
\STATE \quad Update the total cost: $\mathcal{C}_t \leftarrow \mathcal{C}_t + \mathcal{C}(\vec{x}_t,\vec{z}_t)$
\STATE \quad Aggregate data: $\mathcal{D} \leftarrow \mathcal{D} \cup \{(\vec{x}_{t}, \vec{y}_{t},\vec{z}_t)\}$ 
\STATE \quad Update models $\mathcal{GP}_1,\cdots, \mathcal{GP}_K$ 
\STATE \quad $t \leftarrow t+1$
\STATE \textbf{end while}
\STATE \textbf{return} Pareto front and Pareto set of black-box functions $f_1(x), \cdots,f_K(x)$
\end{algorithmic}
\label{alg:CFMESMO}
\end{algorithm}
\end{minipage}
\hfill
\begin{minipage}{0.49\textwidth}
\begin{algorithm}[H]
\centering
\scriptsize
\caption{Naive-CFMO Algorithm}
\textbf{Input}: input space $\mathcal{X}$; $K$ blackbox functions $f_j$ and their continuous approximations $g_j$; total budget $\mathcal{C}_{total}$
\begin{algorithmic}[1] 
\STATE Initialize continuous fidelity gaussian process $\mathcal{GP}_1, \cdots, \mathcal{GP}_K$ by evaluating at initial points $D$
\STATE \textbf{While} {$\mathcal{C}_{t} \leq \mathcal{C}_{total}$} \textbf{do}
 \STATE \quad for each sample $s \in {1,\cdots,S}$: 
 \STATE \quad \quad Sample highest-fidelity functions $\Tilde{f}_j \sim \mathcal{GP}_j(.,z_j^*) $
 \STATE \quad \quad $\mathcal{F}_s^{*} \leftarrow$ Solve {\em cheap} MOO over $(\Tilde{f}_1, \cdots, \Tilde{f}_K)$
 \STATE \quad Find the query based on $\mathcal{F}^{*}=\{\mathcal{F}_s^{*}, s \in \{1 \dots S\}\}$:
\textcolor{red}{ \STATE \quad // Use eq \ref{MESMO} for $\alpha_t$ (MESMO)
 \STATE \quad  select $\vec{x}_{t} \leftarrow \arg max_{\vec{x}\in \mathcal{X}} \hspace{2 mm} \alpha_{t}(\vec{x},\mathcal{F}^{*}) $
\STATE \quad \textbf{for } {$j \in {1 \cdots K}$} \textbf{do}
\STATE \quad \quad select $z_{j} \leftarrow  \arg min_{\vec{z_j}\in \mathcal{Z}_t^{(j)}(\vec{x_t}) \cup \{z_j^*\} } \hspace{2 mm} \mathcal{C}_i(x_t,z_j)  $ \vspace{0.9 mm}
\STATE \quad  Fidelity vector $\vec{z}_{t} \leftarrow [z_1 \dots z_K]$}
 \STATE \quad Update the total cost: $\mathcal{C}_t \leftarrow \mathcal{C}_t + \mathcal{C}(\vec{x}_t,\vec{z}_t)$ 
\STATE \quad Aggregate data: $\mathcal{D} \leftarrow \mathcal{D} \cup \{(\vec{x}_{t}, \vec{y}_{t},\vec{z}_t)\}$
\STATE \quad Update models $\mathcal{GP}_1,\cdots, \mathcal{GP}_K$ 
\STATE \quad $t \leftarrow t+1$
\STATE \textbf{end while}
\STATE \textbf{return} Pareto front and Pareto set of black-box functions $f_1(x), \cdots,f_K(x)$
\end{algorithmic}
\label{alg:CFMESMONaive}
\end{algorithm}
\end{minipage}

%% file: Experiments.tex
\section{Experiments and Results}
In our experiments, we employed CF-GP models as described in section \ref{surrogatesection} with squared exponential kernels. We initialize the surrogate models of all functions with same number of points selected randomly from both lower and higher fidelities.

 We compare iMOCA with several baselines:  six state-of-the-art single-fidelity MO algorithms (ParEGO, SMSego, EHI, SUR, PESMO, and MESMO), one naive continuous-fidelity baseline that we proposed in Section \ref{cf-naive-section}. For experiments in discrete fidelity setting, the number of fidelities is very limited. Thus, the fidelity space reduction method deem meaningless in this case. Therefore, we employ iMOCA without the fidelity space reduction. Additionally, we  compare to the state-of-the-art discrete fidelity method MF-OSEMO. MF-OSEMO has two variants: MF-OSEMO-TG and MF-OSEMO-NI. Since MF-OSEMO-TG has the same formulation as iMOCA-T and provide similar results, we compare only to MF-OSEMO-NI. 

\subsection{Synthetic Benchmarks}
\begin{figure*}[h!] 
    \centering
    \begin{minipage}{0.5\textwidth}
    \centering
    \begin{minipage}{0.49\textwidth}
        \centering
        \includegraphics[width=0.89\textwidth]{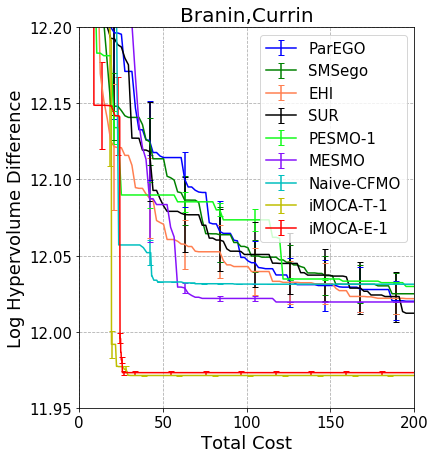} 
    \end{minipage}\hfill
    \begin{minipage}{0.49\textwidth}
        \centering
        \includegraphics[width=0.86\textwidth]{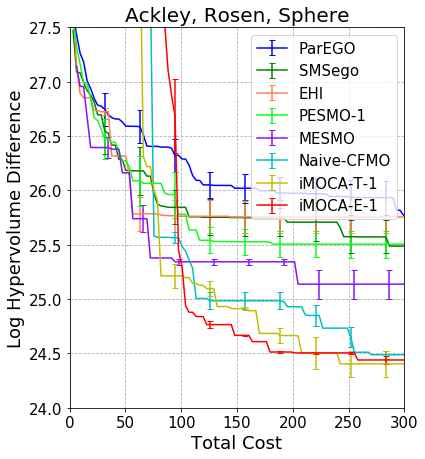} 
    \end{minipage}
    \end{minipage}\hfill
    \begin{minipage}{0.5\textwidth}
            \centering
    \begin{minipage}{0.49\textwidth}
        \centering
        \includegraphics[width=0.85\textwidth]{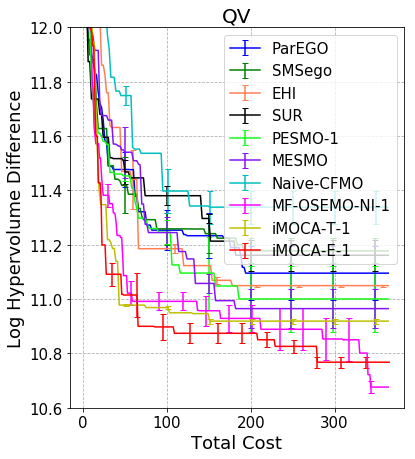} 
    \end{minipage}\hfill
    \begin{minipage}{0.49\textwidth}
        \centering
        \includegraphics[width=0.84\textwidth]{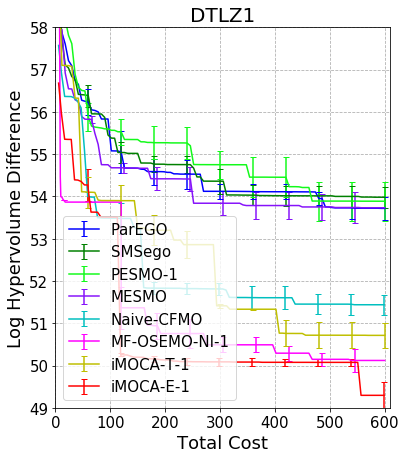} 
    \end{minipage}
    \end{minipage}
          \begin{minipage}{0.5\textwidth}
            \centering
    \begin{minipage}{0.49\textwidth}
        \centering
        \includegraphics[width=0.83\textwidth]{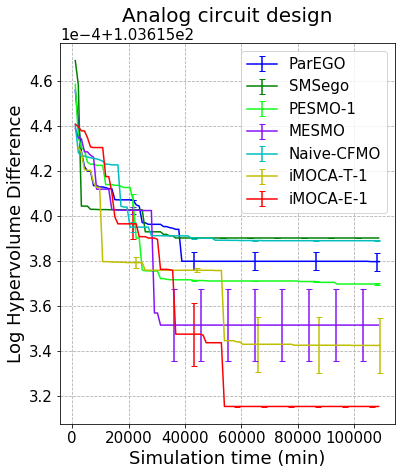} 
    \end{minipage}\hfill
    \begin{minipage}{0.49\textwidth}
        \centering
        \includegraphics[width=0.84\textwidth]{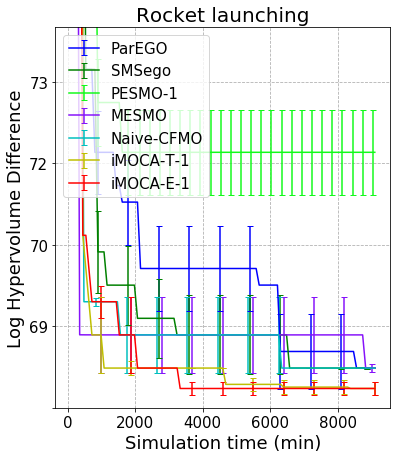} 
    \end{minipage}
    \end{minipage}\hfill
      \begin{minipage}{0.5\textwidth}
    \centering

        \begin{minipage}{0.49\textwidth}
        \centering
        \includegraphics[width=0.86\textwidth]{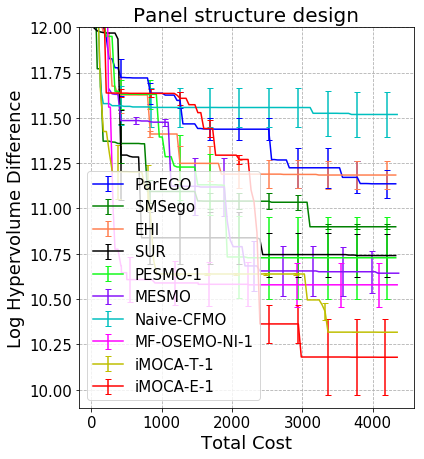} 
    \end{minipage}
    \hfill
    \begin{minipage}{0.49\textwidth}
        \centering
        \includegraphics[width=0.83\textwidth]{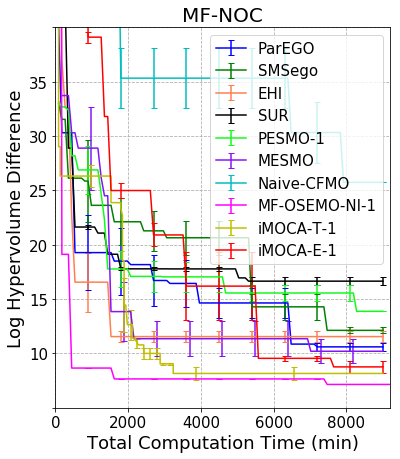} 
    \end{minipage}
    \end{minipage}
\caption{Results of iMOCA and the baselines algorithms on synthetic benchmarks and real-world problems. The $PHV$ metric is presented against the total resource cost of function evaluations.} 
\label{syntheticexp}
\end{figure*}

We evaluate our algorithm iMOCA and baselines on four different synthetic benchmarks.  We construct two problems using a combination of 
benchmark functions for continuous-fidelity and single-objective optimization (\cite{simulationlib}): \emph{Branin,Currin} (with $K$=2, $d$=2) and \emph{Ackley, Rosen, Sphere} (with $K$=3, $d$=5). To show the effectiveness of iMOCA on settings with discrete fidelities, we employ two of the known general MO benchmarks: \emph{QV} (with $K$=2, $d$=8) and \emph{DTLZ1} (with $K$=6, $d$=5) (\cite{habib2019multiple,shu2018line}). Due to lack of space, we provide their complete details in Appendix \ref{sec:appSynthetic}. The titles of plots in Fig. \ref{syntheticexp}, Fig. \ref{syntheticexpsample}, and Fig. \ref{syntheticexpR2} denote the corresponding experiments. 
\subsection{Real-world Engineering Design Optimization Problems} 
We evaluate iMOCA and baselines on four real-world design optimization problems from diverse engineering domains. We provide the details of these problems below. 

\vspace{0.8ex}

{\bf Analog circuit design optimization.} Design of a 
voltage regulator via Cadence circuit simulator that imitate the real hardware (\cite{hong2019dual}).  
Each candidate circuit design is defined by 33 input variables ($d$=33). We optimize nine objectives: efficiency, four output voltages, and four output ripples. This problem has a continuous-fidelity space with cost varying from 10 mins to 120 mins. 

\vspace{0.8ex}

{\bf Panel structure design for large vessels.} The deck structure in large vessels commonly require the design of panels  
resisting uni-axial compression in the direction of the stiffeners (\cite{zhu2014multi}). We consider optimizing the trade-off between two objective functions: weight and strength of the panel. These functions depend on six input variables ($d$=6): one of them is the number of stiffeners used and five others relating to the plate thickness and stiffener dimensions. This problem has a discrete fidelity setting: two fidelities with computational costs 1 min and 21 mins respectively.  

\vspace{0.8ex}

{\bf Rocket launching simulation.} Rocket launching studies (\cite{hasbun2012classical}) require several long and computationally-expensive simulations to reach an optimal design. In this problem, we have three input variables ($d=3$): mass of fuel, launch height, and launch angle. 
The three objective functions are return time, angular distance, and difference between the launch angle and the radius at the point of launch. The simulator has a parameter that can be adjusted to perform continuous fidelity simulations.
We employ the parameter range to vary the cost from 0.05 to 30 mins.

\vspace{0.8ex}

{\bf Network-on-chip.} Communication infrastructure is critical for efficient data movement in hardware chips and they are designed using cycle-accurate simulators. We consider a dataset of 1024 configurations of a network-on-chip with ten input variables ($d$=10) (\cite{rodinia-benchmark}). 
We optimize two objectives: latency and energy. This problem has two discrete fidelities with costs 3 mins and 45 mins respectively. 

\subsection{Results and Analysis}
We compare iMOCA with both approximations (iMOCA-T and iMOCA-E) to all baselines. We employ two known metrics for evaluating the quality of a given Pareto front: {\em Pareto hypervolume ($PHV$)} metric and $R_2$ indicator. 
$PHV$ (\cite{zitzler1999evolutionary}) is defined as the volume between a reference point and the given Pareto front; and $R_2$ is a distance-based metric defined as the average distance between two Pareto-fronts.
We report both the difference in the hyper-volume, and the average distance between an optimal Pareto front $(\mathcal{F^{*}}) $ and the best recovered Pareto front estimated by optimizing the posterior mean of the models at the highest fidelities (\cite{PESMO}). The mean and variance of $PHV$ and $R_2$ metrics across 10 different runs are reported as a function of the total cost.

Fig. \ref{syntheticexp} shows the $PHV$ results of all the baselines and iMOCA for synthetic and real-world experiments (Fig. \ref{syntheticexpR2} in Appendix \ref{addtional_res} shows the corresponding $R_2$ results). We observe that: 1) iMOCA consistently outperforms all baselines. Both iMOCA-T and iMOCA-E have significantly lower converge cost. 2) iMOCA-E shows a better convergence rate than iMOCA-T. This result can be explained by its tighter approximation. Nevertheless, iMOCA-T displays very close or sometimes better results than iMOCA-E. This demonstrates that even with loose approximation, using the iMOCA-T approximation can provide consistently competitive results using less computation time. 3) For experiments with the discrete fidelity setting, iMOCA most of the times outperformed MF-OSEMO or produced very close results. It is important to note that MF-OSEMO is an algorithm designed specifically for the discrete-fidelity setting. Therefore, the competitive performance of iMOCA shows its effectiveness and generalisability.

Figure \ref{syntheticexpsample} in appendix \ref{addtional_res} shows the results of evaluating iMOCA and PESMO with varying number of Monte-Carlo samples $S \in \{1,10,100\}$. For ease of comparison and readability, we present these results in two different figures side by side. We observe that the convergence rate of PESMO is dramatically affected by the number of MC samples $S$. However, iMOCA-T and iMOCA-E maintain a better performance consistently even with a single sample. These results strongly demonstrate that our method {\em iMOCA is much more robust to the number of Monte-Carlo samples}.

\begin{table}[h!]
\centering
\caption{
{\em Best} convergence cost from all baselines $\mathcal{C}_B$, {\em Worst} convergence cost for iMOCA $\mathcal{C}$, and cost reduction factor $\mathcal{G}$.
}\label{tab:costreduction}
\resizebox{0.5\linewidth}{!}{
\begin{tabular}{lllll}  
\toprule
Name & BC  & ARS   & Circuit  & Rocket   \\
\midrule
$\mathcal{C}_B$ & 200  & 300   & 115000 & 9500  \\
\midrule
$\mathcal{C}$ & 30  & 100  & 55000 & 2000  \\
\midrule
$\mathcal{G}$ & 85\%  & 66.6\%  & 52.1\% & 78.9\% \\
\bottomrule 
\end{tabular}
}
\end{table}
\noindent {\bf Cost reduction factor.} 
We also provide the \textit{cost reduction factor} for experiments with continuous fidelities, which is the percentage of gain in the convergence cost when compared to the best performing baseline (the earliest cost for which any of the single-fidelity baselines converge). Although this metric gives advantage to baselines, the results in Table \ref{tab:costreduction} show a consistently high gain ranging from $52.1\%$ to $85\%$. 

%% file: appendix.tex
\section{Full derivation of acquisition function}

 Our goal is to derive a full approximation for iMOCA algorithm. In this appendix, we provide the technical details of the extended-skew Gaussian approximation (iMOCA-E) for the computation of the information gain per unit cost.

 The information gain in equation \ref{af:def} is defined as the expected reduction in entropy $H(.)$ of the posterior distribution $P(\mathcal{F}^* | D)$ due to evaluating $\vec{x}$ at fidelity vector $\vec{z}$. Based on the symmetric property of information gain, we can rewrite it as shown below:

\begin{align}
    I(\{\vec{x}, \vec{y},\vec{z}\}, \mathcal{F}^{*} | D) &= H(\vec{y} | D, \vec{x},\vec{z}) - \mathbb{E}_{\mathcal{F}^{*}} [H(\vec{y} | D,   \vec{x},\vec{z}, \mathcal{F}^{*})] \numberthis \label{eqn_symmetric_igA}
\end{align}
 In equation \ref{eqn_symmetric_igA}, the first term is the entropy of a $K$-dimensional Gaussian distribution  that can be computed in closed form as follows:
\begin{align}
H(\vec{y} | D, \vec{x},\vec{z}) = \sum_{j = 1}^K \ln (\sqrt{2\pi e} ~ \sigma_{g_j}(\vec{x},z_j)) \numberthis \label{firstpartA}
\end{align}
The second term of equation \ref{eqn_symmetric_igA} is an expectation over the Pareto front of the highest fidelities $\mathcal{F}^{*}$. This term can be approximated using Monte-Carlo sampling: 
\begin{align}
    \mathbb{E}_{\mathcal{F}^{*}} [H(\vec{y} | D,   \vec{x},\vec{z}, \mathcal{F}^{*})] \simeq \frac{1}{S} \sum_{s = 1}^S [H(\vec{y} | D,   \vec{x},\vec{z}, \mathcal{F}^{*}_s)] \label{eqn_summationA}
\end{align}
In the main paper, we showed that :
\begin{align}
 y_j &\leq f_s^{j*} \quad \forall j \in \{1,\cdots,K\} \label{inequalityA}
\end{align}
By combining the inequality \ref{inequalityA} and the fact that each function is modeled as an independent CF-GP, a common property of entropy measure allows us to decompose the entropy of a set of independent variables into a sum over entropies of individual variables \cite{information_theory}:
\begin{align}
H(\vec{y} | D,   \vec{x},\vec{z}, \mathcal{F}^{*}_s) \simeq \sum_{j=1}^K H(y_j|D, \vec{x},z_j,f_s^{j*}) \label{eqn_sep_ineqA}
\end{align} 
In what follows, we provide details of iMOCA-E approximation to compute $H(y_j|D, \vec{x},z_j,f_s^{j*})$.

The condition $y_j \leq f_s^{j*}$, is originally expressed as $f_j \leq f_s^{j*}$. Substituting this condition with it's original equivalent, the entropy becomes $H(y_j|D, \vec{x},z_j,f_j\leq f_s^{j*}) $. Since $y_j$ is an evaluation of the function $g_j$ and $f_j$ is an evaluation of the function $f_j$, we make the observation that $y_j | f_j \leq f_s^{j*}$ follows an extended-skew Gaussian (ESG) distribution \cite{azzalini1985class}. It had been shown that the differential entropy of an ESG does not have a closed-form expression \cite{arellano2013shannon}. Therefore, we derive a simplified expression where most of the terms are analytical by manipulating the components of the entropy as shown below. 

In order to simplify the calculation $H(y_j|D, \vec{x},z_j,f_j\leq f_s^{j*})$, we start by deriving an expression for its probability distribution. Based on the definition of the conditional distribution of a bi-variate normal, $f_j | y_j$ is normally distributed with mean $\mu_{f_j}+ \frac{\sigma_{f_j}}{\sigma_{g_j}}\tau(y_j-\mu_{g_j})$ and variance $\sigma_{f_j}^2(1-\tau)^2$, where $\tau= \frac{\sigma_{g_j,f_j}^2}{\sigma_{g_j}\sigma_{f_j}}$ is the predictive correlation between $y_j$ and $f_j$. We can now write the cumulative distribution function for $y_j|f_j\leq f_s^{j*}$ as shown below:
\begin{align*}
    &P(y_j\leq u | f_j \leq f_s^{j*}) = \frac{P(y_j\leq u , f_j \leq f_s^{j*})}{P( f_j \leq f_s^{j*})} = \frac{\int_{-\infty}^u \phi \left(\frac{\theta - \mu_{g_j}}{\sigma_{g_j}}\right) \Phi \left( \frac{f_s^{j*}-\mu_{f_j}- \frac{\sigma_{f_j}}{\sigma_{g_j}}\tau(\theta-\mu_{g_j})}{\sqrt{\sigma_{f_j}^2(1-\tau)^2}} \right) d\theta}{\sigma_{g_j} \Phi \left(\frac{f_s^{j*}-\mu_{f_j}}{\sigma_{f_j}}\right)}
\end{align*}
Let us define the normalized variable $\Gamma_{f_s^{j*}}$ as $\Gamma_{f_s^{j*}}= \frac{y_j - \mu_{g_j}}{\gamma_{g_j}} |f_j\leq f_s^{j*} $. After differentiating with respect to $u$, we can express the probability density function for $\Gamma_{f_s^{j*}}$ as: 
\begin{align*}
    P(u)=\frac{\phi(u)}{\Phi(\gamma_s^{(f_j)})}\Phi(\frac{\gamma_s^{(f_j)}-\tau u}{\sqrt{1-\tau^2}})
\end{align*}
which is the density of an ESG with mean and variance defined as follows: 
\begin{align}
   & \mu_{\Gamma_{f_s^{j*}}}=\tau \frac{\phi(\gamma_s^{(f_j)})}{\Phi(\gamma_s^{(f_j)})},  \sigma_{\Gamma_{f_s^{j*}}}=1- \tau^2\frac{\phi(\gamma_s^{(f_j)})}{\Phi(\gamma_s^{(f_j)})}\left[\gamma_s^{(f_j)}+\frac{\phi(\gamma_s^{(f_j)})}{\Phi(\gamma_s^{(f_j)})} \right]\label{momentsesg}
\end{align}
Therefore, we can express the entropy of the ESG as shown below:
\begin{align}
    H(\Gamma_{f_s^{j*}})= - \int P(u) \ln(P(u))du
\end{align}
We also derive a more simplified expression of the iMOCA-E acquisition function based on ESG.  For a fixed sample $f_s{^j*}$, $H(\Gamma_{f_s^{j*}})$ can be decomposed as follows: 
\begin{align}
    H(\Gamma_{f_s^{j*}})&=\mathbb{E}_{u \sim \Gamma_{f_s^{j*}}}\left[ -\ln(\phi(u)) + \ln(\Phi(\gamma_s^{(f_j)})) -\ln(\Phi(\frac{\gamma_s^{(f_j)}-\tau u}{\sqrt{1-\tau^2}}))\right] \numberthis \label{eq1A}
\end{align}
We expand the first term as shown below:
\begin{align}
    \mathbb{E}_{u \sim \Gamma_{f_s^{j*}}}\left[ -\ln(\phi(u))\right]=\frac{1}{2} \ln(2\pi) +\frac{1}{2}  \mathbb{E}_{u \sim \Gamma_{f_s^{j*}}}\left[ u^2\right]
\end{align}
From the mean and variance of $\Gamma_{f_s^{j*}}$ in equation \ref{momentsesg}, we get: 
\begin{align}
    \mathbb{E}_{u \sim \Gamma_{f_s^{j*}}}\left[ u^2\right] &= \mu_{\Gamma_{f_s^{j*}}}^2 +  \sigma_{\Gamma_{f_s^{j*}}}=1- \tau^2\frac{\phi(\gamma_s^{(f_j)})\gamma_s^{(f_j)}}{\Phi(\gamma_s^{(f_j)})}
\end{align}
We note that the final entropy can be computed using the following expression.
\begin{align}
    H(y_j|D, \vec{x},z_j,y_j\leq f_s^{j*})=  H(\Gamma_{f_s^{j*}})+ \ln(\sigma_{g_j}) \label{ESappriximationA}
\end{align}
By combining equations \ref{eq1A} and \ref{ESappriximationA}, we get:
\begin{align}
    H(y_j|D, \vec{x},z_j,f_j\leq f_s^{j*}) &= \ln(\sqrt{2\pi e} ~\sigma_{g_j}) +\ln(\Phi(\gamma_s^{(f_j)})) - \tau^2\frac{\phi(\gamma_s^{(f_j)})\gamma_s^{(f_j)}}{2\Phi(\gamma_s^{(f_j)})}\nonumber \\
    & \quad - \mathbb{E}_{u \sim \Gamma_{f_s^{j*}}}\left[   \ln(\Phi(\frac{\gamma_s^{(f_j)}-\tau u}{\sqrt{1-\tau^2}}))\right] \numberthis \label{entropyapprox2app}
\end{align}

From equations \ref{eqn_summationA}, \ref{firstpartA}, and \ref{entropyapprox2app}, the final expression of iMOCA-E can be expressed as follows:
\begin{align*}
    \alpha_t(\vec{x},\vec{z},\mathcal{F}^{*})=&\frac{1}{\mathcal{C}(\vec{x},\vec{z})S}\sum_{j=1}^K \sum_{s=1}^S \tau^2\frac{\gamma_s^{(f_j)}\phi(\gamma_s^{(f_j)})}{2\Phi(\gamma_s^{(f_j)})} - \ln(\Phi(\gamma_s^{(f_j)})) +\mathbb{E}_{u \sim \Gamma_{f_s^{j*}}}[\ln(\Phi(\frac{\gamma_s^{(f_j)}-\tau u}{\sqrt{1-\tau^2}}))] 
\end{align*}
Since the differential entropy of an ESG cannot be computed analytically, we
perform numerical integration via Simpson’s rule using $\mu_{\Gamma_{f_s^{j*}}} \mp \gamma \sqrt{\sigma_{\Gamma_{f_s^{j*}}}} $ as the integral limits. In practice, we set $\gamma$ to 5. Since this integral is over one-dimension variable, numerical integration can result in a tight approximation with small amount of computation.

\section{Additional Experiments and Results}

\subsection{Description of Synthetic Benchmarks}
\label{sec:appSynthetic}

In what follows, we provide complete details of the synthetic benchmarks employed in this paper. Since our algorithm is designed for maximization settings, we provide the benchmarks in their maximization form. 

\subsubsection*{1) Branin, Currin experiment}

In this experiment, we construct a multi-objective problem using a combination of existing single-objective optimization benchmarks \cite{kandasamy2017multi}. It has two functions with two dimensions ($K$=2 and $d$=2). 

\textbf{Branin function:}
We use the following function where $\mathcal{C}(z) = 0.05 + z^{6.5}$ 

$$ g(\vec{x},z) = - \left(a(x_2 - b(z)x_1^2 + c(z)x_1 - r)^2 + s(1-t(z))cos(x_1) + s \right)$$

where $a = 1$, $b(z)=5.1/(4\pi^2) - 0.01(1-z)$,
$c(z) = 5/\pi - 0.1(1-z)$, $r=6$, $s=10$ and $t(z)=1/(8\pi) + 0.05(1-z)$.

\textbf{Currin exponential function:}
We use $\mathcal{C}(z) = 0.1 + z^2$
\begin{align*}
g(\vec{x},z) &=- \left(1-0.1(1-z)\exp\left(\frac{-1}{2x_2}\right)\right)
  \left(\frac{2300x_1^3 + 1900x_1^2 + 2092x_1 + 60}{100x_1^3 + 
  500x_1^2 + 4x_1 + 20}\right).
\end{align*}

\subsubsection*{2) Ackley, Rosen, Sphere  experiment}

In this experiment, we construct a multi-objective problem using a combination of existing single-objective optimization benchmarks \cite{wu2018continuous}. It has three functions with five dimensions ($K$=3 and $d$=5). For all functions, we employed $\mathcal{C}(z) = 0.05 + z^{6.5}$ 

\textbf{Ackley function} 

$$ g(\vec{x},z)= -\left(-20 \exp \left[-0.2{\sqrt {\frac{1}{d}\sum_{i=1}^d x_i^{2}}}\right] -\exp \left[\frac{1}{d}\sum_{i=1}^d \cos (2\pi x_i)\right]+e+20\right)-0.01(1-z)$$

\textbf{Rosenbrock function:}

$$ g(\vec{x},z)=-\sum _{i=1}^{d-1}\left[100\left(x_{i+1}-x_{i}^{2}+0.01(1-z)\right)^{2}+\left(1-x_{i}\right)^{2}\right]$$

\textbf{Sphere function:}
$$g(\vec{x},z)=-\sum _{i=1}^{d}x_{i}^{2}-0.01(1-z)$$

\subsubsection*{3) DTLZ1 experiment}

In this experiment, we solve a problem from the general multi-objective optimization benchmarks \cite{habib2019multiple}. We have six functions with five dimensions ($K$=6 and $d$=5) with a discrete fidelity setting. Each function has three fidelities in which $z$ takes three values from $\{0.2,0.6,1\}$ with $z^*$=1. The cost of evaluating each fidelity function is $\mathcal{C}(z)$=$\{0.01,0.1,1\}$

$$g_j(\vec{x},z)=f_j(\vec{x})-e(\vec{x},z)$$

$f_1(\vec{x})=-(1+r)0.5\Pi_{i=1}^5 x_i$

$f_j(\vec{x})=-(1+r)0.5(1-x_{6-j+1})\Pi_{i=1}^{6-j} x_i$ with $j= 2 \dots 5$

$f_6(\vec{x})=-(1+r)0.5(1-x_{1})$

$r=100[d+\sum_{i=1}^d((x_i-0.5)^2)-cos(10 \pi (x_i-0.5))]$

$e(\vec{x},z)=\sum_{i=1}^d \alpha(z)cos(10 \pi \alpha(z)x_i+0.5 \pi \alpha(z)+\pi)$ 
 with $\alpha(z)=1-z$ 
 
\subsubsection*{4) QV experiment}

In this experiment, we solve a problem from the general multi-objective optimization benchmarks \cite{shu2018line}. We have two functions with eight dimensions ($K$=2 and $d$=8) with a discrete fidelity setting.

\textbf{Function 1} has only one fidelity which is the highest fidelity
$$ f_1(\vec{x})=-(\frac{1}{d}\sum_{i=1}^d(x_i^2-20 \pi x_i+10))^{\frac{1}{4}}  $$

\textbf{Function 2} has two fidelities with cost $\{0.1,1\}$ respectively and the following expressions: 

High fidelity: $ f_2(\vec{x}, High)=-(\frac{1}{d}\sum_{i=1}^d((x_i-1.5)^2-20 \pi (x_i-1.5)+10))^{\frac{1}{4}}  $

Low fidelity: $f_2(\vec{x}, Low)=-(\frac{1}{d}((\sum_{i=1}^d(\vec{\alpha[i]}(x_i-1.5)^2-20 \pi (x_i-1.5)+10))^{\frac{1}{4}}  $

with $\vec{\alpha}$=$[0.9,1.1,0.9,1.1,0.9,1.1,0.9,1.1]$

\subsection{Additional results}\label{addtional_res}
\begin{figure*}[h!] 
    \centering
    \begin{minipage}{0.5\textwidth}
    \centering
    \begin{minipage}{0.49\textwidth}
        \centering
        \includegraphics[width=0.89\textwidth]{figures/bc_phv_1_new_new.png} 
    \end{minipage}\hfill
    \begin{minipage}{0.49\textwidth}
        \centering
        \includegraphics[width=0.89\textwidth]{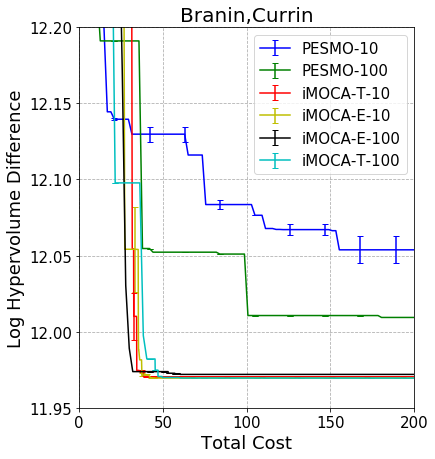} 
    \end{minipage}
    \end{minipage}\hfill
    \begin{minipage}{0.5\textwidth}
            \centering
    \begin{minipage}{0.49\textwidth}
        \centering
        \includegraphics[width=0.85\textwidth]{figures/ARS_phv1_new_new.png} 
    \end{minipage}\hfill
    \begin{minipage}{0.49\textwidth}
        \centering
        \includegraphics[width=0.85\textwidth]{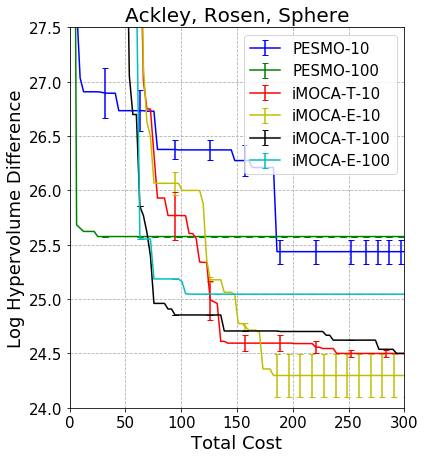} 
    \end{minipage}
    \end{minipage}
        
\caption{Results of synthetic benchmarks showing the effect of varying the number of Monte-Carlo samples for iMOCA, MESMO, and PESMO. The hypervolume difference is shown against the total resource cost of function evaluations.}
\label{syntheticexpsample}
\end{figure*}

\begin{figure*}[h!] 
    \centering
    \begin{minipage}{0.5\textwidth}
    \centering
    \begin{minipage}{0.49\textwidth}
        \centering
        \includegraphics[width=0.89\textwidth]{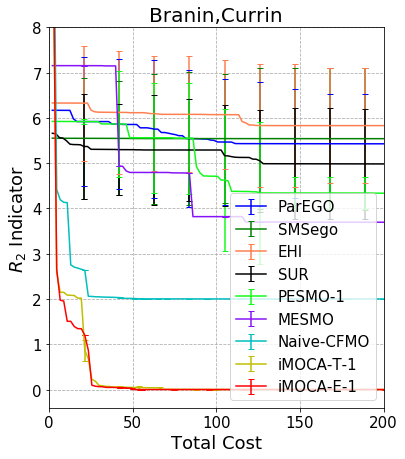} 
    \end{minipage}\hfill
    \begin{minipage}{0.49\textwidth}
        \centering
        \includegraphics[width=0.89\textwidth]{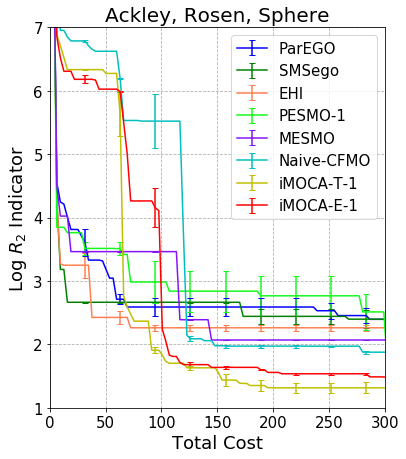} 
    \end{minipage}
    \end{minipage}\hfill
    \begin{minipage}{0.5\textwidth}
            \centering
                \begin{minipage}{0.49\textwidth}
        \centering
        \includegraphics[width=0.85\textwidth]{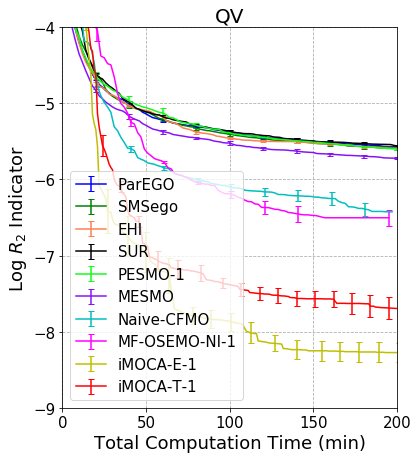} 
    \end{minipage}\hfill
    \begin{minipage}{0.49\textwidth}
        \centering
        \includegraphics[width=0.85\textwidth]{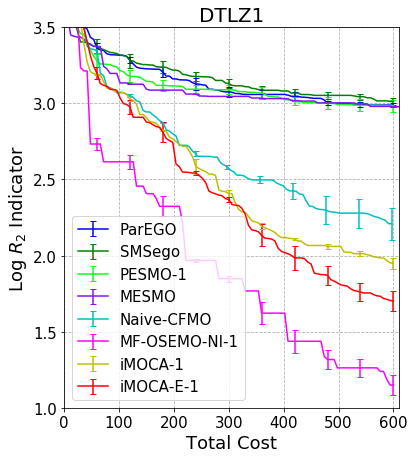} 
    \end{minipage}

    \end{minipage}
          \begin{minipage}{0.5\textwidth}
            \centering
    \begin{minipage}{0.49\textwidth}
        \centering
        \includegraphics[width=0.85\textwidth]{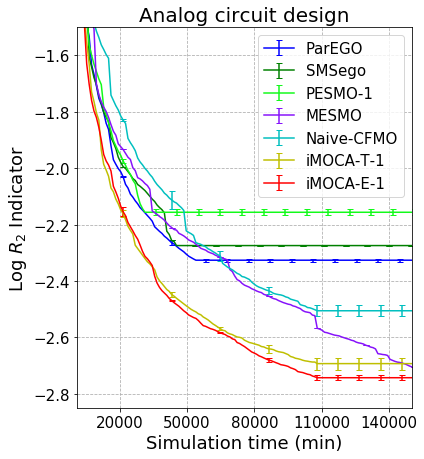} 
    \end{minipage}\hfill
    \begin{minipage}{0.49\textwidth}
        \centering
        \includegraphics[width=0.83\textwidth]{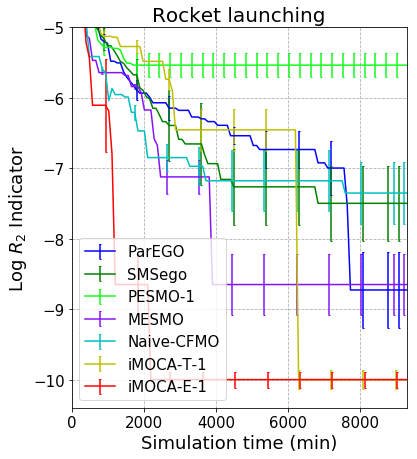} 
    \end{minipage}
    \end{minipage}\hfill
      \begin{minipage}{0.5\textwidth}
    \centering
    \begin{minipage}{0.49\textwidth}
        \centering
        \includegraphics[width=0.85\textwidth]{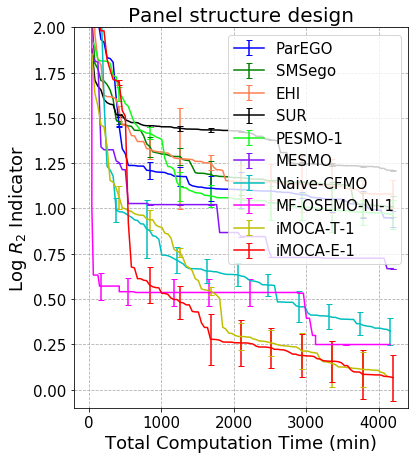} 
    \end{minipage}\hfill
    \begin{minipage}{0.49\textwidth}
        \centering
        \includegraphics[width=0.83\textwidth]{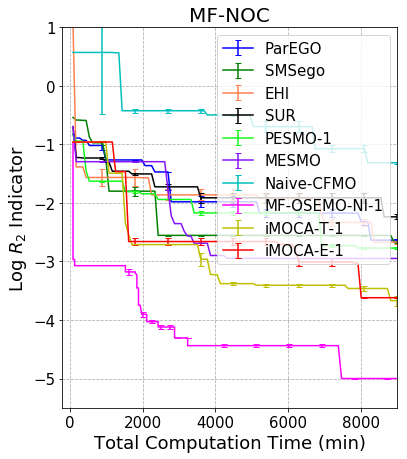} 
    \end{minipage}
    \end{minipage}
    
\caption{Results of iMOCA and the baselines algorithms on synthetic benchmarks and real-world problems. The $R_2$ metric is presented against the total resource cost of function evaluations.}
\label{syntheticexpR2}
\end{figure*}